\documentclass[11pt,a4paper]{article}
\usepackage[hyperref]{eacl2021}
\usepackage{times}
\usepackage{latexsym}

\usepackage{graphicx}
\usepackage{svg}
\usepackage{microtype}
\usepackage{amsmath}
\usepackage{array}
\usepackage{subfig}
\usepackage{wrapfig}
\usepackage{float}
\usepackage{lipsum,  babel}
\usepackage{multirow}
\usepackage[export]{adjustbox}
\usepackage{wrapfig}
\usepackage[ruled,vlined]{algorithm2e}
\usepackage{caption}[2007/04/11]
\aclfinalcopy 
\def\aclpaperid{6} 

\graphicspath{{images/}}

\title{hBert + BiasCorp - Fighting Racism on the Web}

\author{Olawale Onabola\(^1\), Zhuang Ma\(^2\), \\\textbf{Yang Xie\(^3\), Benjamin Akera\(^1\), Abdulrahman Ibraheem\(^1\)},\\\textbf{Jia Xue\(^4\), Dianbo Liu\(^1\), and Yoshua Bengio\(^{1,}\)\(^5\)} \\  \(^1\)Montreal Institute for Learning Algorithms (Mila), \(^2\)Carnegie Mellon University, CMU, \\  \(^3\)Independent Researcher,  \(^4\)University of Toronto, \\ \(^5\)CIFAR Program Co-director \\ {\tt walexi4great@gmail.com, liudianbo@gmail.com} \\ {\tt yoshua.bengio@mila.quebec}}

\begin{document}
\maketitle
\begin{abstract}



Subtle and overt racism is still present both in physical and online communities today and has impacted many lives in different segments of the society.
In this short piece of work, we present how we're tackling this societal issue with Natural Language Processing.
We are releasing BiasCorp \footnote{corresponds to:{dianbo.liu/ yoshua.bengio}@mila.quebec}\footnote{https://doi.org/10.7910/DVN/KPBRLC}, a dataset containing 139,090 comments and news segment from three specific sources -  Fox News, BreitbartNews and YouTube. The first batch (45,000 manually annotated) is ready for publication. We are currently in the final phase of manually labelling the remaining dataset using Amazon Mechanical Turk.
BERT has been used widely in several downstream tasks. In this work, we present hBERT, where we modify certain layers of the pretrained BERT model with the new Hopfield Layer. hBert generalizes well across different distributions with the added advantage of a reduced model complexity.
We are also releasing a JavaScript library and a Chrome Extension Application \footnote{https://mila.quebec/en/project/biasly}, to help developers make use of our trained model in web applications (say chat application) and for users to identify and report racially biased contents on the web respectively.
\end{abstract}

\section{Introduction}
The internet has evolved to become one of the main sources of textual information for many people. Through social media, reviews, and comment sections across the internet, people are continuously consuming information through text. With this, racially biased content has become more entrenched within the language of the internet. Racially biased content in this context refers to the attitudes or stereotypes expressed against marginalized races. 
This is often as a result of implicit bias resulting into hate speech. In this work, we attempt to automatically detect this racially biased content from data collected from the web, including comments from online news outlets such as Fox News and and comments from YouTube videos. We label this dataset  with pointers to racial bias and use machine learning techniques to automate this task. Specifically, we implement BERT as a base model to do this. We also implement a browser extension as a tool to help people identify racially biased content in the information they are consuming. We will also be releasing our curated dataset - BiasCorp to allow more research to be done in this direction.

\section{Related works}
One of the earliest papers to investigate machine learning approaches for the automatic detection of racially-biased online content is \cite{greevy_2004_princip}. The paper identified the potential use of bag-of-words, n-grams, and distributions of parts-of-speech tags as features for the task. Their bag-of-words features are informed by ideas from the field of information retrieval, and involve either word frequencies or counts of word occurrences.  Using an SVM classifier, for bag-of-words features, they found that the use of frequency of words, rather than number of occurrence of words, yielded greater classification accuracies. The n-grams and parts-of-speech tags techniques were unavailable as of the time of their writing.  

 In \cite{warner}, authors followed the definition of \cite{nockleby} by defining hate speech as “any communication that disparages a person or a group on the basis of some characteristic such as race, color, ethnicity, gender, sexual orientation, nationality, religion, or other characteristic.” Their work focused more on detecting anti-Semitic hate speech. For their work, they created a dataset containing hate speech obtained from Yahoo! and the American Jewish Congress. Following the work of \cite{yarowsky_1994}, they employed hand-crafted template-based features. Apart from the fact that these features are hand-engineered, a potential drawback is their sheer size: a total of 3,537 features, which is prone to the curse of dimensionality. A counter-intuitive result reported by the paper is that the uni-gram features contributed best to classification accuracies. They used linear-kernel SVMs for classification.
 
 The work of \cite{yahoo_nobata} dealt with the broad category of abusive language. Authors of the work gave definitions for distinguishing between three categories of abusive language: hate speech which subsumes racial bias, derogatory remarks and profanity. Further, they described reasons why automatic detection of abusive language, which subsumes racial bias, is difficult. Reasons include: clever evasion of detection engines by users via the use of mischievous permutations of words (e.g. Niggah written as Ni99ah); evolution of ethnic slurs with time; role of cultural context in the perception and interpretation of slurs, as a phrase that is considered derogative in one culture might be perfectly neutral in another culture. Towards building their classification model, they employed four categories of features namely, n-grams, lexical features, syntactic/parser features, and word-level as well as comment-level embeddings. They found that character-level n-grams gave the highest contribution to the model’s accuracy.  
 
The authors of \cite{us_and_them} also developed techniques for detecting multiple hate speech categories including the racially-based category. Towards creating their datasets, they harnessed hate speech event-triggers. For example, to create their racial bias dataset, they collected tweets in a two-week interval following the re-election of Barrack Obama as U.S president.  They explored a number of potential features towards building their classification algorithm: bag of words, lexicon of hateful terms, and typed dependencies. In addition, they experimented into classification via SVMs versus classification via random forests, and reported that the former yielded superior performance over the latter. Also, they compared the use of classifiers trained for each hate speech category against the use of a single classifier trained on data spanning all categories. As expected, the specialized classifiers outperformed their multi-category counterpart. 
\cite{demo_embed} followed the definition of \cite{gelber_2007}, which states that hate speech is: ”speech or expression which is capable of instilling or inciting hatred of, or prejudice towards, a person or group of people on a specified ground, including race, nationality, ethnicity, country of origin, ethno-religious identity, religion, sexuality, gender identity or gender.” The main research thrust of their work was to apply demographic embeddings \cite{bamman} , \cite{hovy}, for the task of racial bias detection in tweets. Compared to other works such as \cite{us_and_them},  for instance, a particularly distinguishing result of \cite{demo_embed} is how their data extraction procedure is able to arrive at a better balanced ratio of racially-biased to non-racially-biased comments. For example, in the work, 40.58 percent of Canadian tweets were judged racially-biased by human annotators, whereas in \cite{us_and_them} only about 3.73 percent of the comments in the dataset are racially biased. Classification results using an SVM classifier revealed benefits of their proposed  demographic embeddings over traditional features and embeddings.
In \cite{saleh_2020_white_sup}, the authors explored the detection of hate speech in White supremacist forums. They explored BiLSTM, logistic regression and BERT for their task. Also, they compared the use of domain-agnostic pre-trained word embedding (such as GloVe.6B.300d ) versus the use of a domain-aware 300-dimensional word2vec embedding trained on the specific dataset used in the work. Results showed that BERT yields better results than both logistic regression and BiLSTM. Further, results proved  the domain-aware embeddings to be superior to the pre-trained embeddings.

\section{Method}
\subsection{Data curation and processing}
The datasets used for training were obtained from discussion channels of online news media by programmed web crawler based on Scrapy framework with all crawled data stored in PostgreSQL database. Since existing comments of online article were generally loaded by asynchronous API accessed by a specific key hidden in the articles before presenting them on website, the web crawler parsed keys for each article after completing a list with URLs of all articles waiting to be further crawled and then matched the keys with their corresponding API to retrieved stored comments for each article. 

First, sentences containing neural racial words from a curated list were selected. Second, the sentiment score of each comment was calculated according to two lookup tables: a combined and augmented \cite{jockers_2017} and Rinker's augmented Hu and Liu \cite{sentimentr} \cite{hu_liu_aug_sentimentr} positive/negative word list as sentiment lookup values, and a racial-related English lookup table from Hatebase\footnote{https://hatebase.org/}. To guarantee these two tables influence the sentiment score consistently, the lookup values of the Hatebase table were adjusted by percentage. Then we extracted the data with bottom 20 percent of the sentiment score, and matched them up with other randomly selected comments appearing under the same articles or videos as random control. Finally, equal numbers of random controls are added into the data set, to ensure that approximately half of the data is racially discriminatory.


\subsection{Model Architecture}
Attention-based Transformer network \cite{vaswani2017attention} has been used widely across different natural language processing tasks. Based on the previous successes of the transformer network, we decided to use the BERT Architecture \cite{devlin2019bert} as our base model. Unlike previous variant of the attention-based language models such as \cite{gpt}, BERT learns to jointly conditions on the right and left context of the input representation at all the layers by randomly masking out segments of the input token. This is particularly useful for extracting contextual information from the input representation, and it's very applicable to our use case. We aim to build a variant of the model that can generalize \textit{sufficiently} well across different data distributions\footnote{distributions here implies different use cases or data environments/sources}. The notion of \textit{sufficiency} is evaluated by training, validating and testing our model on data across the different sources. We fine-tune the pretrained BERT model on our curated dataset rather than training from scratch (this choice was based on empirical results).
We are releasing a JavaScript library for developers to use our pretrained model in front facing applications such as chat app, to flag down racially biased comments. Consequently, we need to optimize for the model complexity without sacrificing performance gain. BERT has a huge number of parameters / large model size. Other methods have been employed to reduce the complexity without hurting the performance, such as knowledge distillation \cite{distillbert} and quantization \cite{bert_quant}. It has also been proven that pruning the weights of the pretrained model do not necessarily affect the model performance, within acceptable 'thresholds' \cite{gordon2020compressing}. In a similar fashion, we aim to reduce the complexity of BERT without sacrificing performance by replacing certain layers with the Hopfield layer \cite{ramsauer2020hopfield}. 
Hopfield layer can be used to replace the attention-based layer of the BERT model; as it has been shown to approximate the functionality of the attention mechanism with a new Energy update rule (modified version of the Hopfield network extended to continuous state representation). The learning dynamics of BERT as shown in \cite{ramsauer2020hopfield} shows that the attention heads in the higher layers are mostly responsible for extracting task-specific features from the input representation. We replaced the self-attention mechanism in the last \( X \) layers of the pretrained BERT model with a Hopfield layer, where \( X \) is an hyperparameter. In a similar approach described in \cite{vaswani2017attention}, we use residual connection around the Hopfield sub-layer, followed by layer normalization \cite{ba2016layer}. It has been shown that residual connections help propagate positional information across layers.
The replaced Hopfield layer drastically reduced the parameter size of our model. To further improve the performance of the model, we use the Hopfield Pooling layer which acts as both a permutation equivariant layer and pools generated embedding from the modified BERT model. The Hopfield pooling layer also acts as a form of memory to store the hidden state of the last layer in the modified BERT model.
Finally, we add a classification layer on top of the pooling layer for the task in question. 
\subsection{Model Training}
\begin{figure*}[h!]
\centering
\includegraphics[width=\textwidth]{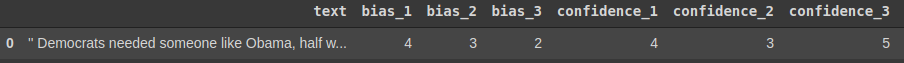} 
\caption{Sample annotation}
\label{fig:sample}
\end{figure*}

\begin{figure*}[h!]
\centering
    \subfloat[Bias Score]{\includegraphics[width=0.5\textwidth]{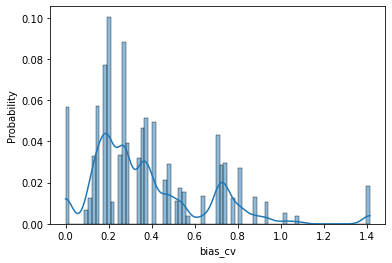} }
    \subfloat[Confidence Score]{\includegraphics[width=0.5\textwidth]{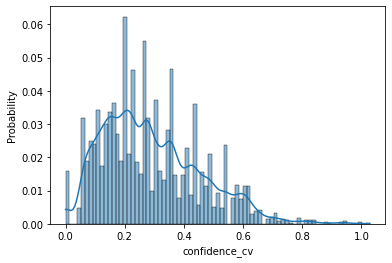}}
    \caption{Confidence of Variation}
    \label{fig:cov}
\end{figure*}

Given the disparity between the annotators for each sample in our dataset, averaging the labels with the confidence scores as weights might be noisy.
We computed the coefficient of variation \(CV\) among annotators for each sample in our dataset.
Using the recommended \cite{cov_article} \cite{DBLP:journals/corr/VeitACKGB17} \(CV\) of \textit{0.2} for the bias scores would imply dropping \(90\%\) of the dataset as seen in \ref{fig:cov}. In order to fully utilize the dataset and effectively manage the disparity between the annotators, we formulate a loss function \(\mathcal{L}_{model}\) given by
\begin{equation}
    L_{model} = 1/N\sum_{i=1}^{N} CE\bigg(p\big(x_i\big),q\big(x_i\big)\bigg)
\end{equation}
where \(CE\big(p\big(x_i\big),q\big(x_i\big)\big)\) is the cross entropy between \(p(x_i)\) and \(q(x_i)\) for the \(ith\) sample, and \(N\) is the size of the dataset.
\begin{equation}
    CE(p,q) = -\sum_{i=1}^{c}p_c(x)\log(\epsilon + q_c(x))
\end{equation}
\(q_c(x)\) is the predicted probability of sample \(x\) in class \(c\), equivalently, the output probabilities from the model and \(\epsilon\) is for numerical stability. \(p_c(x)\) is the probability of sample \(x\) in class \(c\), equivalently, \(p_c(x)\) is a \(c-length\) vector with entries such that \(\sum_{i=1}^{c}p_c(x)=1\). The entries of \(p_c(x)\) are the normalized confidence scores of the annotators with indices given by the respective voted classes. As an example, following the algorithm described in \ref{algo:form}, for a given sample shown in figure \ref{fig:sample}; the bias scores of the \(3\) different annotators with their confidence level is represented with an array of tuples, \(X\) where each tuple,  \((b_i,s_i)\) is the bias score \(b_i\) with the associated confidence score, \(s_i\) by annotator \(i\). To calculate \(p_c(x)\), we first normalize the confidence scores across the \(3\) different annotators such that \(\sum_{i=1}^{3}s_i=1\). The resulting \(p_c(x)\) for the entry, \(S\), shown in \ref{fig:sample} is
\begin{align*}
X &= \bigg[ (4,4), (3,3), (2,5) \bigg] \\
X_{norm} &=  \bigg[ (4,0.3333), (3,0.25), (2,0.4167) \bigg] \\
p_c(X) &= [ 0., 0., 0.4167, 0.25, 0.3333, 0. ]
\end{align*}
\normalsize
\begin{algorithm}
\SetAlgoLined\SetArgSty{2em}
\KwResult{\(p_c(x)\) }
\BlankLine
 \textbf{\emph{Input:}} An array of target scores \textbf{\emph{t}}, and array of confidence scores
 \textbf{\emph{s}} where \textbf{\emph{s[i]}} is the confidence score by \textbf{\emph{ annotator i}} for choosing target score \textbf{\emph{t[i]}} \\ Both arrays are of equal length N where N is the number of annotators. \textbf{\emph{C}} is the number of classes (equivalently the range/max of possible target scores if scores are integer.) \\
\BlankLine
\textbf{\emph{Step 1: Initialize \(p_c \leftarrow [\:\:.0 \quad \textbf{\emph{for}}\quad \textbf{\emph{\_}} \quad \textbf{\emph{in}} \quad C]\)}}\\
\BlankLine
\textbf{\emph{Step 2: Calculate normalizing constant K}}\\
\BlankLine
\textbf{ \(K \leftarrow \sum_{i=1}^{N} \textbf{\emph{$s_i$}} \) } \;
\BlankLine
\textbf{\emph{Step 3: Set the values of \(p_c\)}}\\
\BlankLine
\For{i in N}{
  \(  class\_index \leftarrow t[i] \)\;
  \(p_c[ class\_index ] \xleftarrow{+} \frac{s[i]}{K}\) \;
 }
 \caption{Compute \(p_c(x)\) for a sample x}
 \label{algo:form}
  \end{algorithm}
 
\subsection{Evaluation Task and Metrics}
We evaluate the model performance across the validation and test set, given that they are from different distributions or sources. The test set contains only comments from YouTube while the validation set was randomly sampled from Fox News and BreitbartNews. The particular choices were due to the fact that the first batch of the dataset used for training contained very relatively few samples from YouTube.
We evaluate our approach using two methods; multiclass classification and multiclass-multilabel classification. \\
\textbf{Using the multiclass approach,} for a given sample, \( k \) and using the method described previously in calculating the target class, the class with the maximum confidence score was used as the target. We calculate the average precision for each class, \( AP_c \) and the mean average precision  \( MAP\) averaged over the entire dataset with size \( N \) along the class dimension \( d \) as described in \cite{DBLP:journals/corr/VeitACKGB17}
\begin{align} \label{eq:metrics}
AP_c &=  \frac{\sum_{k=1}^{N} Precision(k, c) \cdot rel(k,c)}{number of positives} \\ 
MAP &=  1/d\sum_{c=1}^{d} AP_c
\end{align}
\onecolumn
\begin{figure*}[h!]
    \subfloat[Train Vs Validation Loss]{\includegraphics[width=0.9\textwidth, height=0.3\textwidth]{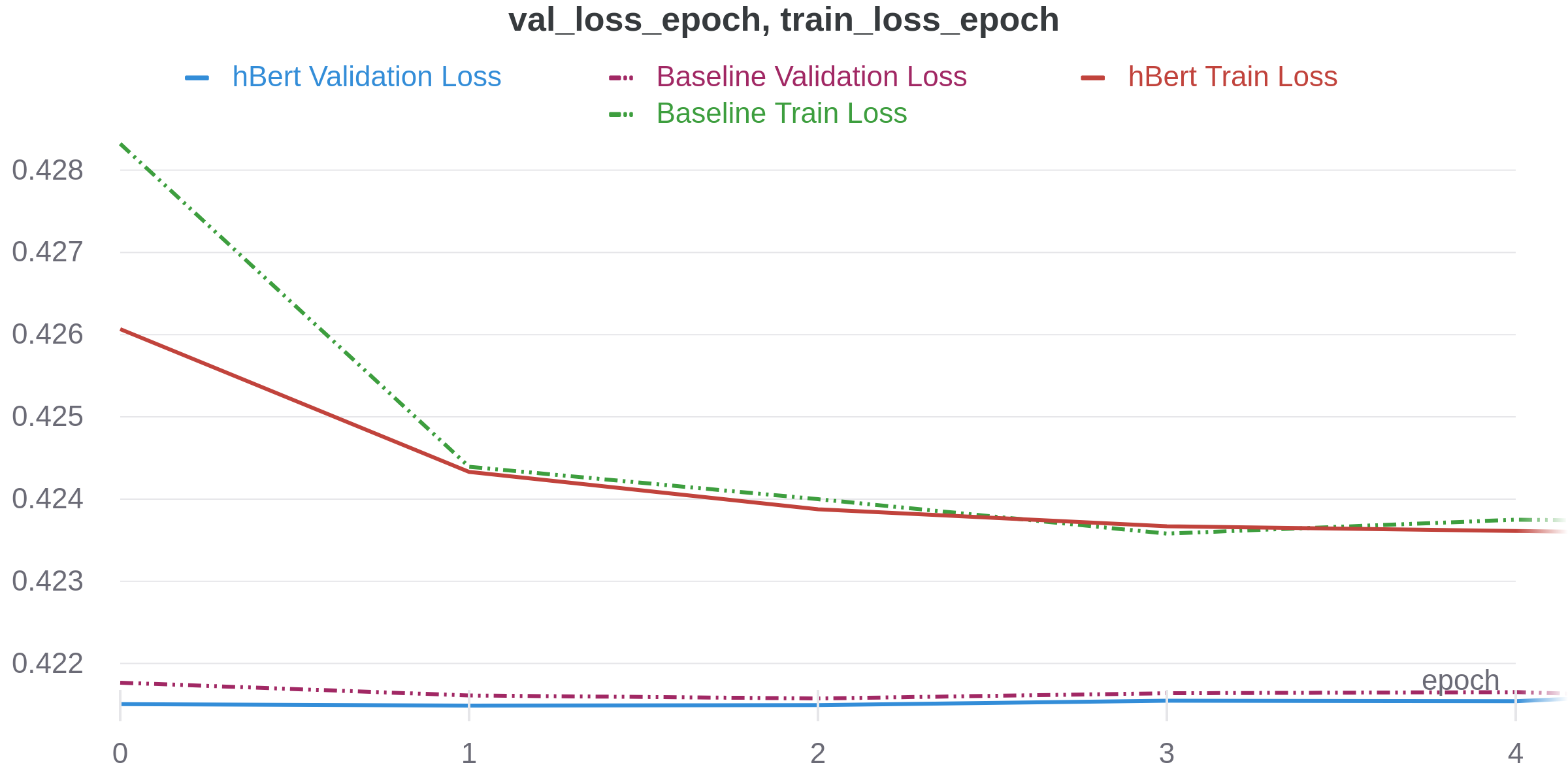} }
    \label{fig:trainloss}
\end{figure*}

\begin{table*}[h!]
\centering
\resizebox{\textwidth}{!}{
\begin{tabular}{|l|l|l|l|l|l|l|l|l|l|l|}
    \hline
    \multirow{2}{*}{Model} &
    \multicolumn{3}{c|}{TopK Accuracy} &
    \multirow{1}{*}{mAP}  &
    \multicolumn{3}{c|}{F1 @ k}  &
    \multicolumn{3}{c|}{IoU @ k} \\
    & 1 & 2 & 3 &  & 1 & 2 & 3 & 1 & 2 & 3\\
    \hline
    Baseline & 0.6015625 & 0.703125 & 0.7890625 & 0.29355 & 0.5859 & 0.6953 & 0.7734 & 0.2102 & 0.2114 & 0.2102\\
    \hline
    hBert & \textbf{0.640625} & 0.703125 & 0.765625 & \textbf{0.3501} & \textbf{0.6562} & \textbf{0.7109} & \textbf{0.8125} & \textbf{0.2266} & \textbf{0.2165} & \textbf{0.2281} \\
    \hline
  \end{tabular}
  }
  \caption{Test Metrics for selected trial for each model configuration}
  \label{tab:summary1}
\end{table*}
\begin{table*}[h!]
\centering
\resizebox{0.6\textwidth}{!}{
\begin{tabular}{|l|l|l|l|l|l|l|}
    \hline
    \multirow{2}{*}{Model} &
    \multicolumn{6}{c|}{AP}   \\
    & 0 & 1 & 2 & 3 & 4 & 5\\
    \hline
    Baseline & 0.2205 & 0.0967 & 0.1344 & 0.9564 & 0.1103 &0.2340 \\
    \hline
    hBert & 0.1195 & \textbf{0.1111} & \textbf{0.2132} & \textbf{0.9607} & \textbf{0.5049} & 0.1914 \\
    \hline
\end{tabular}
  }
  \caption{The Average Precision (AP) for the different classes}
  \label{tab:summary2}
\end{table*}
\begin{figure*}[tb!]
        \centering
    \includegraphics[width=1\textwidth, height=25em]{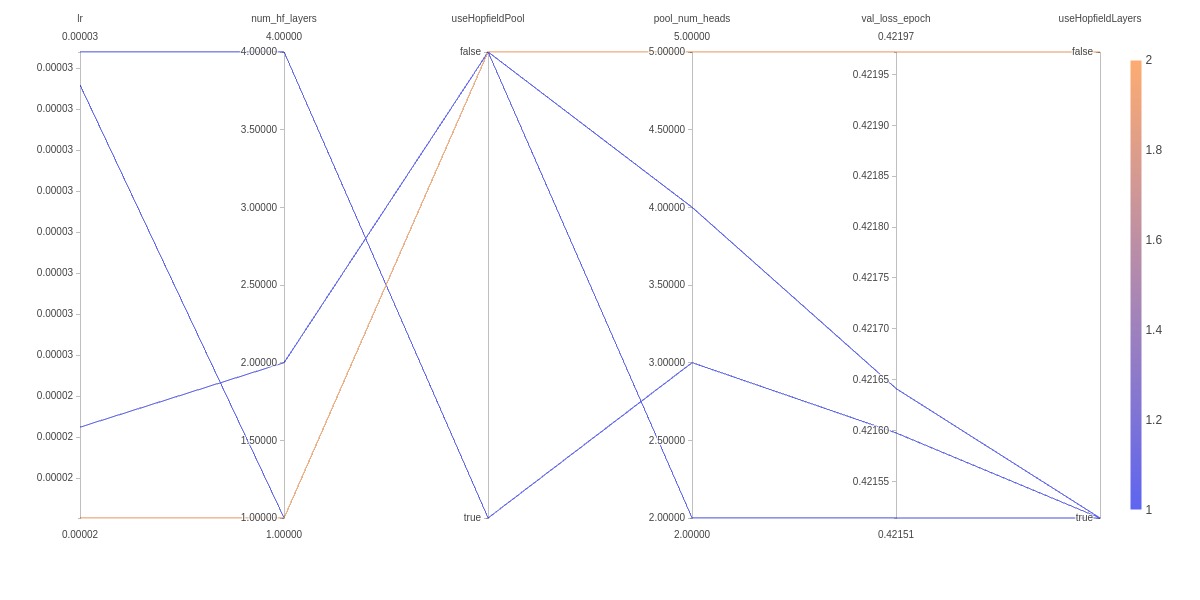}
\captionsetup{justification=centering}
\caption{Parallel Coordinate Graph for multiple runs/trials across model configurations \\ The model configuration is the Baseline when the target variable (\textbf{\emph{useHopfieldLayers}} in the graph is False. \\ The \textbf{\emph{useHopfieldPool}} variable denotes whether the Hopfield Pooling layer was used \\
The \textbf{\emph{ lr, pool\_num\_heads, num\_hf\_layers, val\_loss\_epoch }} variables in the graph are the learning rate, the number of heads in the Hopfield Pooling Layer (if used), the number of Hopfield Layer and the validation loss respectively}. With a reduced model complexity, the hBert performs relatively as good as the baseline
\label{fig:paralel}
\end{figure*}
\twocolumn
where \(Precision(k,c) \) is the precision for class \( c\) for the \(kth\) sample and \( rel(k,c) \) is an indicator function that is 1 if the predicted and the target class for sample \(k\) is positive. We also report the \(topK\) accuracy, for \(k=[1,3]\) since we had a max of 3 annotators for each \(k\). \\
\textbf{Using the multilabel approach,} for a given sample, \( k \) and using the method described previously in calculating the target class, we take the top k classes as the target classes. We do the same for the predictions (obtained after passing the output logits through a softmax function). We compute the \(AP_c\) (for each class), \(mAP, F1\) score, and \(IoU\)
\section{Experiments}
\subsection{Training Details \& Result}
We run a multi-objective hyperparameter search (using Optuna\cite{optuna_2019}) optimizing for the following parameters: validation loss, FLOPs (indicative of the model complexity and ultimately the inference time), mAP on the validation and test set, and the Intersection over Union IoU scores (also known as the Jaccard Index) for the \(topk\) for \(k=[1,3]\) transformations described above. We use 4 NVidia V100SXM2 (16G memory) GPUs on a single node, with batch size of 32. We reduced the batch size (instead of say 64) because we had to run multiple trials and to avoid the notorious OOM error. For each model configuration, we run 10 trials with 5 epochs each.
As seen in \ref{fig:paralel}, the hBert perform relatively better with a reduced model complexity. In \ref{tab:summary1}, the models predictions were more accurate for an increasing \(k\). The hBert perform better than the Baseline for the Top1 accuracy. The F1 scores and Jaccard Index (IoU) for the hBert were relatively higher for \(k=[1,3]\). The \(mAP\), which is the average of the \(AP_c\) over the classes, is relatively low because of the low performing classes as seen in \ref{tab:summary2}

\subsection{Data statistics}
The data set contains 139,090 rows, and 67.70 percent of their sentiment scores are negative. Their average sentiment score is -0.1422, and the median value is -0.1203, ranging from -3.6206 to 2.1414.  66,998 of them are comments from Fox News, with an average sentiment score of -0.0997 and a median of -0.0884, ranging from -2.8591 to 2.1414. And 63,948 of the data are comments from Breitbart News, with an average sentiment score of -0.1760 and a median of -0.1721, ranging from -3.6206 to 1.3576. And 8,144 of the data are comments from YouTube, with an average sentiment score of -0.2259 and a median of -0.2694, ranging from -3.3000 to 1.4673.
In this work, we used the first batch of the dataset; which have been manually annotated using Amazon Mechanical Turk. After pre-processing the input text (removing irrelevant tokens such as mentions), the maximum length was 478 (it was 623 before preprocessing).

\section{Discussion}
In this work we have shown a way to detect racial bias in text. We experimented with a BERT-based model as we aim to reduce model complexity without sacrificing much of the performance.
We also discussed the BiasCorp, a manually labelled dataset containing racially biased comments from Fox News, BreitbartNews and YouTube. To enable developers make use of our pretrained hBERT model, we are releasing a Javascript Library, optimized for inference on the edge. A Chrome Extension will also be available for users to help report and identify racially bias text on the web. We also plan to extend this work to other forms of biases such as Gender. In a future work, we plan to further reduce the model complexity by using Gaussian Kernel as described in \cite{ramsauer2020hopfield} and other quantization tricks.
 

\section*{Acknowledgments}
This research was enabled in part by support provided by Calcul Québec (www.calculquebec.ca) and Compute Canada (www.computecanada.ca)

\bibliography{anthology,eacl2021}
\bibliographystyle{acl_natbib}
\appendix
\end{document}